\documentclass[letterpaper,10pt,conference]{ieeeconf}

\IEEEoverridecommandlockouts
\usepackage[top=2cm,bottom=2cm,left=2cm,right=2cm]{geometry}
\usepackage{cite}
\usepackage{amsmath,amssymb,amsfonts}
\usepackage{algorithm}
\usepackage{algpseudocode}
\usepackage{graphicx}
\usepackage{textcomp}
\usepackage{xcolor}
\usepackage{booktabs}
\usepackage{multicol}
\usepackage{multirow}
\usepackage{subcaption}

\begin{document}

\title{Does Visual Information Play a Decisive Role in Vision-Language-Action Model Driving Behavior?}

\author{
Jingtao He, Hongliang Lu, Xiaoyun Qiu, Yixuan Wang, Xinhu Zheng$^{*}$%
\thanks{This paper has been accepted by the 2026 IEEE International Conference on Intelligent Transportation Systems (ITSC 2026).}%
\thanks{The authors are with the Intelligent Transportation Thrust, The Hong Kong University of Science and Technology (Guangzhou), Guangzhou, China.}%
\thanks{$^{*}$Corresponding author: xinhuzheng@hkust-gz.edu.cn}%
\thanks{\copyright~2026 IEEE. Personal use of this material is permitted. Permission from IEEE must be obtained for all other uses, including reprinting, republishing, or reuse of any copyrighted component of this work in other works.}%
}

\maketitle

\begin{abstract}
Vision-Language-Action (VLA) models have demonstrated promising capability in autonomous driving, highlighting the potential of unified multimodal architectures for jointly modeling perception and planning.
However, how current VLA-based driving behavior is grounded in visual information remains poorly understood.
Existing evaluation protocols mainly focus on aggregate performance metrics, lacking structured and practical diagnostics to quantify visual–behavior dependency.
In this work, we introduce a structured multi-level visual perturbation framework to analyze visual–behavior dependency in VLA-based driving models systematically. 
The framework organizes controlled visual perturbations along three complementary dimensions: channel-level degradation, information-level disruption, and structure-level modification. 
We apply it to VLA-based driving systems and evaluate behavioral responses under both open-loop trajectory prediction and interactive closed-loop safety evaluation. 
Experimental results reveal evaluation-dependent dependency patterns and uneven visual grounding across abstraction levels. 
These findings call for more structured analyses and principled design of VLA driving models to better understand how visual information shapes behavior and develop safer, more robust systems.
\end{abstract}


\section{Introduction}
Vision-Language-Action (VLA) models have recently emerged as a promising paradigm for autonomous driving, seeking to integrate visual perception and trajectory planning within a unified multimodal architecture \cite{jiang2025survey, sapkota2025vision}. By directly mapping image observations and auxiliary priors to future trajectories, these models offer an alternative to traditional modular pipelines that explicitly decompose perception, prediction, and planning into separate components. Contemporary studies have shown that such VLA architectures are capable of generating coherent, context-aware driving behaviors across diverse traffic scenarios, highlighting the potential of multimodal end-to-end learning for autonomous driving \cite{Renz2025cvprcarllava,xie2026latentvla,gao2026steervlasteeringvisionlanguageactionmodels}.

Despite these advances, a fundamental question remains insufficiently understood: \textbf{to what extent is driving behavior in VLA systems grounded in visual information?} 
Common evaluation metrics, such as trajectory error, collision rate, and rule compliance, primarily measure behavioral performance under standard input conditions. 
However, strong overall performance does not necessarily imply strong visual grounding, because these metrics do not directly reflect how driving behavior depends on visual input.
In particular, high performance under clean inputs may conceal cases where visual information is only weakly utilized or functionally bypassed, raising concerns about hidden modality imbalance in safety-critical systems.
In safety-critical scenarios, understanding whether decisions are visually grounded is essential for diagnosing failure modes and ensuring reliable interaction in dynamic environments.

In vision–language modeling research, perturbation-based diagnostic techniques have been widely employed to examine modality reliance and shortcut learning behaviors, especially in visual question answering and multimodal reasoning models\cite{selvaraju2019taking,niu2021counterfactual}.
Similar strategies have been explored in autonomous driving, where visual perturbations are primarily studied in the context of robustness evaluation under adverse conditions such as noise, corruption, or weather variation\cite{dong2023benchmarking,10733168}. 
However, robustness-oriented evaluations focus on performance degradation under environmental variation, rather than characterizing how visual information contributes to behavior outcomes. 
In particular, existing studies rarely distinguish between different abstraction levels of visual representation (e.g., sensory statistics, semantic density, or spatial organization), making it difficult to disentangle where and how visual grounding manifests in the perception-to-action pipeline. 
Consequently, the structured visual–behavior dependency in VLA-based autonomous driving systems remains insufficiently characterized.

To address this gap, we introduce a structured diagnostic framework that probes the role of visual grounding in VLA-based autonomous driving models through controlled multi-level perturbations.
Our design organizes visual perturbations across three complementary dimensions: 1) channel-level, which operates at the sensory input stage and alters or removes the visual signal, 2) information-level disruption, which alters semantically salient content, and 3) structure-level modification, which disturbs spatial organization and relational consistency. 
By organizing perturbations along progressively abstract representational levels, we enable a controlled dissection of visual influence across the perception-to-action pipeline. 
We instantiate this framework on a representative VLA driving model and evaluate behavioral responses under both open-loop trajectory prediction and interactive closed-loop evaluation. 
Our empirical analysis reveals heterogeneous, level-dependent behavioral sensitivity: certain degradations induce limited behavioral variation, whereas structured perturbations can significantly influence safety-critical outcomes. 
These results suggest that visual grounding in current VLA systems is context-sensitive and unevenly distributed across perturbation levels. 
Importantly, the contrast between open-loop and closed-loop evaluations indicates that commonly adopted evaluation protocols may provide only a partial view of interaction-critical visual dependency.

Our objective is to provide a systematic empirical characterization of visual–behavior dependency in VLA-based autonomous driving systems. 
By examining behavioral responses under structured perturbation regimes, this study provides a practical diagnostic toolkit for analyzing visual grounding in future VLA-based autonomous driving architectures.

In summary, our contributions are threefold:
\begin{itemize}
\item We propose a structured multi-level visual perturbation framework for systematically analyzing visual–behavior dependency in VLA-based autonomous driving systems.

\item 
We perform systematic evaluations across both open-loop and interactive closed-loop driving settings to assess behavioral sensitivity under controlled visual perturbations.

\item Our empirical findings reveal heterogeneous visual–behavior dependency patterns that vary across evaluation settings in a representative VLA-based driving model.
\end{itemize}

\section{Related Work}
\subsection{Vision-Language-Action Models for Autonomous Driving}
Recent advances in large-scale multimodal learning have enabled the development of VLA models for autonomous driving, which directly map image observations and contextual inputs to future driving actions within unified end-to-end architectures. These systems typically rely on large-scale multimodal pretraining and cross-modal alignment mechanisms to learn shared representations across perception and action. Existing VLA-based driving models differ along several modeling dimensions. From the perspective of action representation, some methods formulate driving behavior as discrete sequential tokens (e.g., AutoVLA \cite{zhou2025autovla}), whereas others predict continuous trajectory distributions (e.g., OpenDriveVLA \cite{zhou2025opendrivevla}). In terms of generative mechanism, approaches range from deterministic regression models to probabilistic and diffusion-style planners (e.g., Recogdrive \cite{li2025recogdrive}). Training strategies further vary between pure imitation learning and hybrid schemes incorporating reinforcement objectives (e.g., Drive-r1 \cite{li2025driver1}, Alpamayo-r1\cite{wang2025Alpamayo-R1}). Although diverse in formulation, these systems consistently aim to integrate scene representation and action generation within multimodal architectures. 
Recent efforts have also explored benchmark design and explainability analysis for Vision-Language models in autonomous driving, aiming to better understand decision-making behaviors under diverse scenarios \cite{tang2026autodridmexplainablebenchmarkdecisionmaking}. While such studies provide valuable insights into model reasoning and failure modes, they do not explicitly analyze how driving behavior depends on structured perturbations in visual input.

\subsection{Modality Dependency and Perturbation-Based Analysis}
Understanding modality-specific dependency has emerged as an important direction in multimodal learning research. Prior studies report that multimodal models may exhibit modality imbalance \cite{peng2022balanced,agrawal2018don}, shortcut learning\cite{geirhos2020shortcut}, or modality collapse, where predictions are disproportionately influenced by certain inputs while others contribute marginally. Such phenomena, observed in vision–language tasks, have motivated closer examination of perceptual grounding and interpretability \cite{goyal2017making}.
Perturbation-based techniques have been widely adopted 
to evaluate model sensitivity under controlled input variations,
particularly in robustness benchmarking \cite{hendrycks2019benchmarking}.
Methods such as token pruning, masking, feature corruption,
and structured degradation enable controlled probing of 
input–output relationships.
However, existing studies primarily focus on performance degradation,
rather than systematically characterizing modality-specific
behavioral dependency patterns.
In autonomous driving, prior studies further suggest that end-to-end models evaluated under open-loop protocols may rely heavily on ego-state or historical trajectory inputs, potentially diminishing apparent sensitivity to visual perturbations \cite{Li2023IsES,tang2025decoupling}. These observations indicate that evaluation settings can substantially influence perceived modality dependency patterns. Nevertheless, in the context of Vision-Language-Action autonomous driving, structured investigations into how driving behavior evolves under progressive and multi-level visual perturbation remain limited.

\section{Methodology}
\begin{figure*}[t]
\centering
\includegraphics[width=\textwidth]{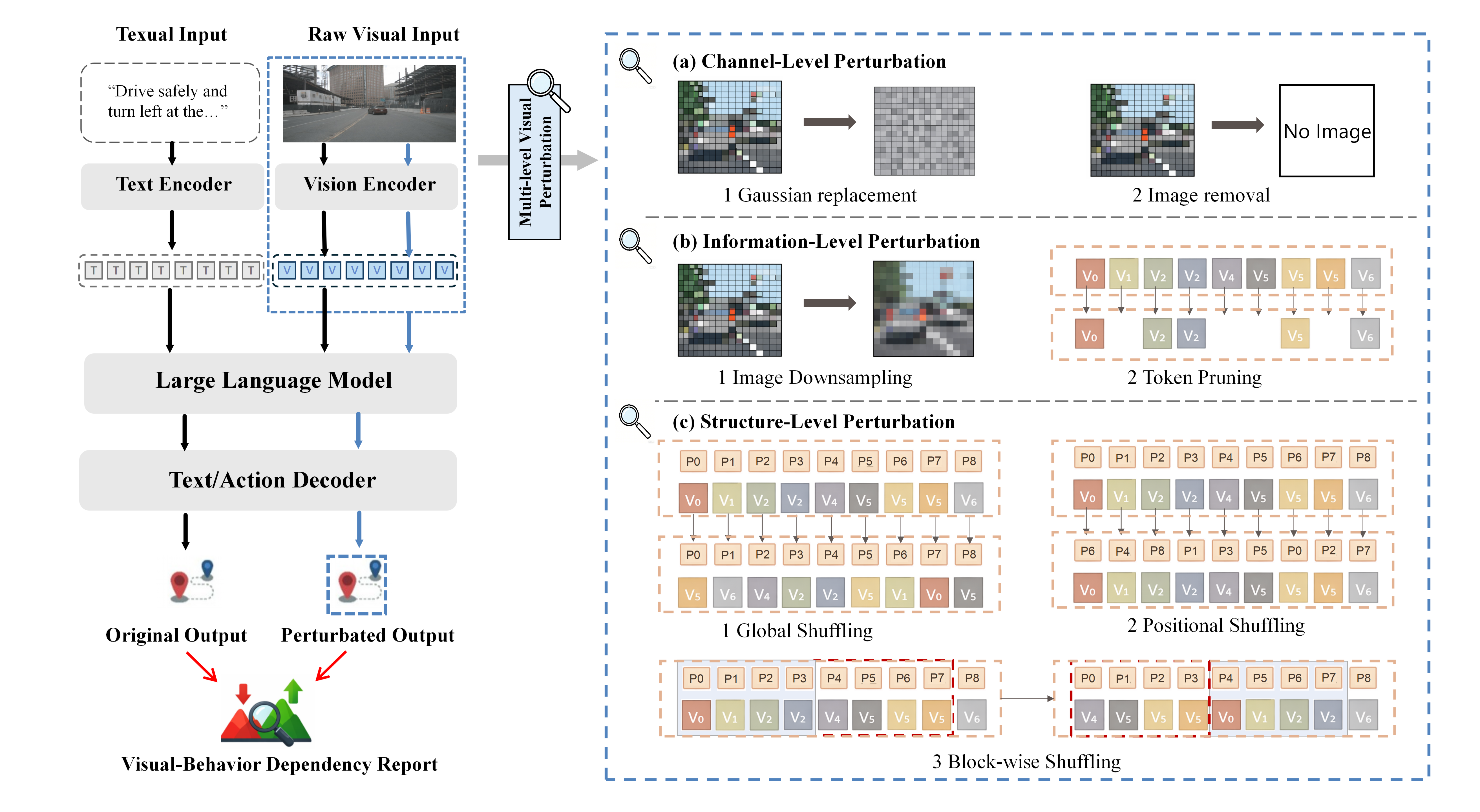}
\caption{Overview of the structured multi-level visual perturbation framework. \textbf{Left:} The VLA processing pipeline mapping multimodal inputs to driving actions.
\textbf{Right:} Three families of visual perturbations applied at channel,
information, and structure levels.}
\label{fig:framework}
\end{figure*}

We introduce a structured multi-level visual perturbation framework
to systematically analyze visual–behavior dependency in VLA-based
autonomous driving systems.
As illustrated in Fig.~\ref{fig:framework}, the framework integrates
the VLA processing pipeline with perturbations organized across
complementary abstraction levels.
This design aligns perturbation granularity with the hierarchical
representation process in VLA models.
Specifically, \textbf{channel-level} perturbations examine the functional role and global robustness of the visual modality by applying
holistic sensory disruptions.
\textbf{Information-level} perturbations assess reliance on semantic content density, while \textbf{structure-level} perturbations probe sensitivity to spatial and relational organization.
Visual–behavior dependency is quantified through performance variation between clean and perturbed visual conditions.

\subsection{Problem Formulation}

We consider a VLA-based driving policy 
$f_{\theta}$ that maps multimodal observations to future driving 
actions or trajectories. At time step $t$, the policy takes a visual observation
 $I_t$ as the primary perceptual input, together with auxiliary modalities $S_t$ (e.g., ego-state signals, historical context, or radar measurements), and produces an action output:
\begin{equation}
a_t = f_{\theta}(I_t, S_t).
\end{equation}

To analyze the dependency between visual perception and action generation, we subject the visual modality to controlled perturbations and examine the resulting behavioral changes.
Let $\mathcal{T}$ denote a visual perturbation operator applied to the visual processing pipeline, producing a perturbed visual representation $\tilde{V}_t$. 
Depending on the level of perturbation, $\tilde{V}_t$ may correspond to either a perturbed image observation or a perturbed encoder representation.
The corresponding policy output is
\begin{equation}
\tilde{a}_t = f_{\theta}(\tilde{V}_t, S_t).
\end{equation}

By comparing behavioral performance under clean and perturbed visual representations, we probe the functional dependency between visual perception and action generation in VLA-based driving systems. 
The formulation is model-agnostic and does not rely on any specific architectural assumptions about $f_{\theta}$.

\subsection{Multi-Level Visual Perturbation Framework}

To systematically analyze visual–behavior dependency, we decompose the visual perturbation operator into three complementary levels reflecting distinct stages of visual processing: channel-level, information-level, and structure-level perturbation.

\paragraph{Multi-level Perturbation}
Let $V_t$ denote the visual representation at the targeted stage,
corresponding to $I_t$ in the image space and to $Z_t = E(I_t)$ in the encoder representation space.
Perturbation is applied at the appropriate stage depending on the targeted aspect of visual processing.

\textbf{Channel-Level Perturbations}
operate directly in the image space and modify low-level sensory signals without altering semantic layout. 
These transformations modify pixel statistics while preserving global scene organization:
\begin{equation}
\tilde{I}_t = \mathcal{T}_{\text{ch}}(I_t).
\end{equation}

Representative operations include noise substitution and complete image removal. 

\textbf{Information-Level Perturbations} target semantic density while maintaining coarse spatial structure. 
Such perturbations can be applied either in the image space or in the encoder representation space:
\begin{equation}
\tilde{V}_t = \mathcal{T}_{\text{inf}}(V_t).
\end{equation}

Representative operations include downsampling followed by upsampling in the image space, and random vision token dropping or structured token pruning in the representation space.

\textbf{Structure-Level Perturbations}
disrupt spatial organization while preserving local visual content. 
This perturbation is implemented in the encoder representation space:
\begin{equation}
\tilde{Z}_t = \mathcal{T}_{\text{str}}(Z_t),
\end{equation}
where $Z_t = E(I_t)$ denotes the encoder token representation.

Representative operations include block-wise vision token shuffling and positional index perturbation. 

\paragraph{Dependency Quantification}

We quantify visual dependency by measuring performance variation under visual perturbations. 
Let $M(\cdot)$ denote a task-specific evaluation metric provided by the benchmark and computed over an evaluation set. 
For a perturbation operator $\mathcal{T}$, we define the relative performance change as

\begin{equation}
D(\mathcal{T}) =
\frac{
M(f_{\theta}(\tilde{V}, S)) -
M(f_{\theta}(V, S))
}{
|M(f_{\theta}(V, S))|
}.
\end{equation}

Here, $V$ and $\tilde{V}$ denote the clean and perturbed visual representations, respectively. 
$D(\mathcal{T})$ captures both the direction and magnitude of
performance change, enabling consistent dependency analysis
across open-loop and closed-loop settings.
Particularly, the sign of $D(T)$ should be interpreted with respect to the definition of the evaluation metric.

\paragraph{Overall Procedure}
Given a selected VLA policy $f_{\theta}$ and a benchmark $\mathcal{B}$, 
the overall execution procedure of the proposed framework is summarized in Algorithm~\ref{alg:framework}.
The metric $M(\cdot)$ is defined by the chosen benchmark 
(e.g., open-loop prediction accuracy or closed-loop safety score),
and is used to quantify policy performance.
The procedure first computes the baseline performance under clean visual input.
It then iterates over perturbations across the channel-level,
information-level, and structure-level families.
For each perturbation, the corresponding perturbed visual representation 
is constructed, the benchmark metric is re-evaluated,
and the relative dependency measure is computed.
This procedure enables systematic comparison of visual–behavior
dependency across different perturbation types and evaluation settings.
\begin{algorithm}[h]
\caption{Multi-Level Visual Perturbation Framework}
\label{alg:framework}
\begin{algorithmic}[1]
\Require VLA policy $f_{\theta}$; benchmark $\mathcal{B}$ with metric function $M(\cdot)$; perturbation family 
         $\mathbb{\mathcal{T}}=\{\mathbb{\mathcal{T}}_{ch},\mathbb{\mathcal{T}}_{inf},\mathbb{\mathcal{T}}_{str}\}$

\State Compute base performance:
\State \hspace{1em} $m_{base} \gets M(f_{\theta}(V,S))$

\For{level $\ell \in \{ch, inf, str\}$}
    \For{$\mathcal{T} \in \mathbb{\mathcal{T}}_{\ell}$}
        \State Construct perturbed visual representation:
        \State \hspace{1em} $\tilde{V} \gets \mathcal{T}(V)$
        \State Evaluate perturbed policy:
        \State \hspace{1em} $m_{\ell}(\mathcal{T}) \gets M(f_{\theta}(\tilde{V}, S))$
        \State Compute relative dependency:
        \State \hspace{1em}
        $D_{\ell}(\mathcal{T})
        \gets
        \dfrac{m_{\ell}(\mathcal{T}) - m_{base}}
        {|m_{base}|}$
    \EndFor
\EndFor
\end{algorithmic}
\end{algorithm}

\section{Experiments}
Having established the structured multi-level perturbation framework, we instantiate it on a representative VLA model and analyze behavioral responses under controlled perturbations.

\subsection{Experimental Setup}
\paragraph{Model}
Impromptu-VLA \cite{chi2025impromptu} is a representative large-scale Vision-Language-Action model built upon the Qwen2.5-VL\cite{qwen2vl2024} backbone and following the unified multimodal paradigm commonly adopted in recent VLA-based autonomous driving research. The model is pretrained on diverse multimodal datasets and further adapted for driving scenarios, providing a broad visual–language foundation for policy generation. We select the publicly released 3B-parameter version due to its open weights, reproducibility, and architectural compatibility with structured perturbation analysis.

\paragraph{Benchmarks}
We adopt \textbf{nuScenes}\cite{caesar2020nuscenes} as a representative and widely used open-loop benchmark for trajectory prediction. 
However, open-loop evaluation primarily measures trajectory accuracy under offline settings and does not incorporate interactive vehicle–environment feedback, motivating complementary closed-loop analysis.

To better analyze safety-critical visual grounding, we additionally perform closed-loop evaluation in the \textbf{NeuroNCAP}\cite{ljungbergh2024neuroncap} simulation environment, which introduces interactive dynamics and explicit safety metrics. 

\paragraph{Metrics} We evaluate the model performance on the selected benchmarks following their respective evaluation protocols.

\textbf{nuScenes (Open-loop).}
Following the Impromptu-VLA evaluation setting, we evaluate on the NavTest split of nuScenes
by computing the $\ell_2$ displacement errors at 1\,s, 2\,s,
and 3\,s horizons and reporting their arithmetic mean (Mean L2 Error (1–3s)), where all distances are measured in meters.

\textbf{NeuroNCAP (Closed-loop).}
We evaluate interactive driving safety using the NeuroNCAP benchmark. Each scenario is executed for 30 runs,
and reported metrics are averaged across runs to reduce stochastic variance.
The following metrics are reported:

(1) \textit{NCAP Score}, the primary safety metric measuring overall driving performance under standardized scenarios.
In the original NeuroNCAP protocol, leaving the drivable area is not explicitly treated as a failure category and may still yield a non-zero score.
To enforce stricter safety constraints, out-of-bounds events are additionally penalized as collision failures in the NCAP scoring, resulting in a zero score for the corresponding episode.
A higher NCAP score indicates safer driving behavior.

(2) \textit{Collision Rate (CR)}, defined as the percentage of scenarios in which a collision occurs.

(3) \textit{Out-of-Bounds Rate (OOB)}, defined as the percentage of episodes where the vehicle leaves the drivable area.

CR and OOB are reported as diagnostic indicators, where lower values indicate safer behavior.

\textbf{Relative Performance Variation}
Relative performance variation $\Delta$ is computed with respect to the clean baseline.
For trajectory evaluation, $\Delta$ is calculated on the Mean L2 Error:
\[
\Delta_{\text{L2}} =
\frac{\text{L2}_{\text{perturbed}} - \text{L2}_{\text{base}}}
{\text{L2}_{\text{base}}}.
\]
For safety evaluation, $\Delta$ is computed on the NCAP score:
\[
\Delta_{\text{NCAP}} =
\frac{\text{NCAP}_{\text{perturbed}} - \text{NCAP}_{\text{base}}}
{\text{NCAP}_{\text{base}}}.
\]

\paragraph{Implementation Details}
All visual perturbations are applied during inference only,
without updating model parameters or retraining.
At each timestep, the model takes as input a single front-view RGB image
together with a fixed textual prompt containing ego-state signals
and historical information from the previous six timesteps (3\,s).
Perturbations are applied independently at each timestep,
ensuring that performance variations reflect instantaneous visual perturbation
rather than accumulated temporal corruption.

\textbf{Channel-level} perturbations replace the original visual input in pixel space while preserving input resolution and model architecture.

(1) \textit{Gaussian replacement(GR)} substitutes the original image with a randomly generated Gaussian replacement image of identical resolution and channel dimension. 
This removes semantic content while preserving low-level channel statistics, thereby isolating the effect of sensory-level degradation on driving behavior.

(2) \textit{Image removal} disables the image channel at inference
by omitting the visual input, effectively reducing the model to
language-conditioned policy generation based solely on ego-state
and historical signals, without modifying the model architecture.

\textbf{Information-level} perturbations reduce visual information density while preserving input resolution and encoder structure.

(1) \textit{Image downsampling} reduces pixel-level detail by
scaling both spatial dimensions by a fixed ratio and then
upsampling back to the original resolution before inference,
thereby reducing effective visual information while preserving
input size.

(2) \textit{Random token pruning} randomly discards a fixed proportion (e.g., 50\%, 75\%, 90\%) of visual tokens during the prefill stage
prior to cross-modal fusion.

(3) \textit{FastV token pruning} removes a fixed proportion
(e.g., 50\%, 75\%, 90\%) of visual tokens following the FastV\cite{chen2024image} strategy,
which selects tokens based on a predefined importance criterion.

In both cases, the pruning ratio is defined with respect to the total
number of visual tokens before fusion, while the remaining encoder
architecture, token dimensionality, and fusion mechanism remain unchanged.

\textbf{Structure-level} perturbations modify the spatial organization of visual tokens while preserving all information, dimensionality, and encoder architecture.

(1) \textit{Global shuffling} randomly permutes all the visual token features within the visual token segment, while keeping their positional encoding indices unchanged. 
Non-visual tokens remain unmodified. 
This operation disrupts the correspondence between visual content and spatial location within the visual modality without altering cross-modal ordering.

(2) \textit{Positional shuffling} randomly permutes the positional encoding indices associated only with the visual token segment under the RoPE-based positional mechanism of Qwen2.5-VL, while keeping the token feature representations and non-visual tokens unchanged. 
This operation alters spatial indexing within the visual modality but preserves visual content and cross-modal structure.

(3) \textit{Block-wise shuffling} groups spatially adjacent visual tokens into predefined blocks (block-4 and block-8) according to their original 2D layout, and then randomly permutes these blocks as whole units. 
Tokens within each block remain unchanged, while block-level relocation disrupts mid-scale spatial organization. 
Non-visual tokens and token count remain unmodified.

\subsection{Channel-Level Perturbation Results}
Table~\ref{tab:channel} reports quantitative results under channel-level perturbations in both open-loop and closed-loop settings.

\begin{table}[h]
\centering
\caption{Channel-level visual perturbation under nuScenes and NeuroNCAP evaluation. $\Delta$ denotes relative change compared to the no perturbation baseline. GR denotes Gaussian replacement. Mean L2 is measured in meters (m); CR and OOB are reported in percent (\%).}
\label{tab:channel}
\setlength{\tabcolsep}{4pt}
\begin{tabular}{lcccc}
\toprule
\multirow{2}{*}{\textbf{Setting}} 
& \multicolumn{1}{c}{\textbf{nuScenes}} 
& \multicolumn{3}{c}{\textbf{NeuroNCAP}} \\
\cmidrule(lr){2-2} \cmidrule(lr){3-5}
& Mean L2 ($\Delta_{\text{L2}}$) 
& NCAP ($\Delta_{\text{NCAP}}$) 
& CR 
& OOB \\
\midrule
No Perturbation 
& \textbf{0.311} 
& \textbf{1.555} 
& 76.6\% 
& 1.5\% \\

GR 
& 0.323 (+3.9\%) 
& 1.471 (-5.4\%) 
& \textbf{73.7\%} 
& 5.4\% \\

Image Removal
& 0.333 (+7.1\%) 
& 1.328 (-14.6\%) 
& 79.3\% 
& \textbf{0.9\%} \\
\bottomrule
\end{tabular}
\end{table}

\textbf{Open-loop Performance.}
Channel-level corruption leads to only limited degradation in open-loop trajectory prediction.
Under Gaussian replacement, the Mean L2 Error increases from 0.311m to 0.323m (+3.9\%), while complete removal of visual input further raises it to 0.333m (+7.1\%).
Notably, even the absence of image features results in less than 10\% relative performance change.
This suggests that offline trajectory estimation relies predominantly on non-visual priors and learned motion patterns rather than fine-grained visual fidelity.

\textbf{Closed-loop Performance.}
Closed-loop safety performance exhibits substantially greater sensitivity to channel-level corruption.
Gaussian replacement reduces the NCAP score from 1.555 to 1.471 (-5.4\%), while complete removal of visual input further degrades it to 1.328 (-14.6\%).
Although collision rate does not increase monotonically, the overall safety score consistently declines as the visual channel is progressively degraded.
Notably, the performance gap between Gaussian corruption and full removal suggests that the mere presence of the visual channel contributes more to safety-critical control than the integrity of its detailed information content.
These results indicate that interactive control depends more critically on semantically grounded visual perception than offline trajectory fitting.

\subsection{Information-Level Perturbation Results}
Table~\ref{tab:info} reports quantitative results under information-level perturbations in both open-loop and closed-loop settings.

\begin{table}[h]
\centering
\caption{Information-level visual perturbation under nuScenes and NeuroNCAP evaluation. $\Delta$ denotes relative change compared to the no perturbation baseline. Mean L2 is measured in meters (m); CR and OOB are reported in percent (\%).}
\label{tab:info}
\setlength{\tabcolsep}{4pt}
\renewcommand{\arraystretch}{1.05}
\begin{tabular}{lcccc}
\toprule
\multirow{2}{*}{\textbf{Setting}} 
& \multicolumn{1}{c}{\textbf{nuScenes}} 
& \multicolumn{3}{c}{\textbf{NeuroNCAP}} \\
\cmidrule(lr){2-2} \cmidrule(lr){3-5}
& Mean L2 ($\Delta_{\text{L2}}$)  
& NCAP ($\Delta_{\text{NCAP}}$) 
& CR 
& OOB \\
\midrule
No Perturbation 
& \textbf{0.311} 
& \textbf{1.555}
& 76.6\% 
& 1.5\% \\

Downsample(50\%)
& 0.316 (+1.6\%) 
& 1.473 (-5.3\%) 
& 78.0\% 
& 0.9\% \\

Downsample(75\%)
& 0.318 (+2.3\%) 
& 1.132 (-27.2\%) 
& 83.9\% 
& 1.9\% \\

Downsample(90\%)
& 0.319 (+2.6\%) 
& 1.066 (-31.5\%) 
& 85.0\% 
& 1.7\% \\

Random(50\%)
& 0.319 (+2.6\%) 
& 1.362 (-12.4\%) 
& 81.0\% 
& \textbf{0.7\%} \\

Random(75\%)
& 0.319 (+2.6\%) 
& 1.397 (-10.2\%) 
& 78.5\% 
& 1.9\% \\

Random(90\%)
& 0.322 (+3.5\%) 
& 1.396 (-10.2\%) 
& 78.8\% 
& 1.4\% \\

FastV(50\%)
& 0.318 (+2.3\%) 
& 1.284 (-17.4\%) 
& 79.0\% 
& 3.5\% \\

FastV(75\%)
& 0.323 (+3.9\%) 
& 1.478 (-5.0\%) 
& 77.9\% 
& 1.0\% \\

FastV(90\%)
& 0.329 (+5.8\%) 
& 1.456 (-6.4\%)
& \textbf{75.5\%} 
& \textbf{0.7\%} \\
\bottomrule
\end{tabular}
\end{table}

\textbf{Open-loop Performance.}
Across all information-level perturbations, open-loop degradation remains limited.
Even under severe reduction settings (up to 90\%), the Mean L2 Error varies within a narrow range (0.311m to 0.329m), corresponding to at most a 5.8\% relative change.
Downsample, Random, and FastV strategies all produce only minor fluctuations in trajectory error.
This indicates that open-loop trajectory metrics are relatively insensitive to large-scale reductions in visual information density.

\textbf{Closed-loop Performance.}
Closed-loop safety is highly sensitive to information-level perturbations.
Aggressive image downsampling severely degrades performance, reducing NCAP by -27.2\% and -31.5\% under 75\% and 90\% downsampling, respectively.
By comparison, pruning encoded visual tokens leads to less pronounced degradation, with the largest drop being -17.4\%.
The larger impact of pre-encoder degradation suggests that semantic formation is more critical than maintaining dense encoded tokens.
These results imply that interactive planning operates on a compact set of high-level semantic representations rather than raw visual richness.

\subsection{Structure-Level Perturbation Results}
Table~\ref{tab:struct} reports quantitative results under structure-level perturbations in both open-loop and closed-loop settings.

\begin{table}[h]
\centering
\caption{Structure-level visual perturbation under nuScenes and NeuroNCAP evaluation. $\Delta$ denotes relative change compared to the no perturbation baseline. Mean L2 is measured in meters (m); CR and OOB are reported in percent (\%).}
\label{tab:struct}
\setlength{\tabcolsep}{4pt}
\renewcommand{\arraystretch}{1.05}
\begin{tabular}{lcccc}
\toprule
\multirow{2}{*}{\textbf{Setting}} 
& \multicolumn{1}{c}{\textbf{nuScenes}} 
& \multicolumn{3}{c}{\textbf{NeuroNCAP}} \\
\cmidrule(lr){2-2} \cmidrule(lr){3-5}
& Mean L2 ($\Delta_{\text{L2}}$)  
& NCAP ($\Delta_{\text{NCAP}}$) 
& CR 
& OOB \\
\midrule
No Perturbation 
&\textbf{0.311} 
& 1.555 
& 76.6\% 
& 1.5\% \\

Global shuffling
& 0.318 (+2.2\%)
& 1.446 (-7.0\%) 
& 78.0\% 
& 1.2\% \\

Positional shuffling
& 0.317 (+1.9\%)
& 1.310 (-15.8\%) 
& 81.4\% 
& \textbf{0.7\%} \\

Block-4
& 0.322 (+3.5\%)
& 1.538 (-1.1\%) 
& 76.4\% 
& 1.7\% \\

Block-8
& 0.325 (+4.5\%)
& \textbf{1.680 }(+8.0\%) 
& \textbf{73.1\%} 
& 1.7\% \\
\bottomrule
\end{tabular}
\end{table}

\textbf{Open-loop Performance.}
Open-loop trajectory prediction remains largely insensitive to structural perturbations.
Across all spatial reorganizations, the Mean L2 Error increases within a narrow range (+1.9\% to +4.5\%), despite disruption of global layout coherence.
This suggests that offline trajectory estimation is relatively insensitive to precise spatial alignment and can tolerate substantial reorganization of visual structure.

\textbf{Closed-loop Performance.}
Closed-loop behavior exhibits scale-dependent sensitivity to structural perturbations.
Global shuffling and positional shuffling significantly degrade safety performance (-7.0\% and -15.8\%, respectively), indicating that large-scale spatial misalignment interferes with coherent scene grounding.
Notably, positional shuffling causes greater degradation than content-only shuffling, highlighting the importance of spatial indexing in visual-language alignment.

In contrast, block-wise shuffling introduces only minor performance variations.
Block-4 results in marginal degradation (-1.1\%).
The improved NCAP score under Block-8 may reflect a shift toward more conservative behavior rather than enhanced scene understanding.
Structural fragmentation could inflate perceived environmental complexity, encouraging risk-averse planning that reduces collision frequency in closed-loop evaluation.

\subsection{Qualitative Case Study}

\begin{figure*}[t]
\centering
\includegraphics[width=\textwidth]{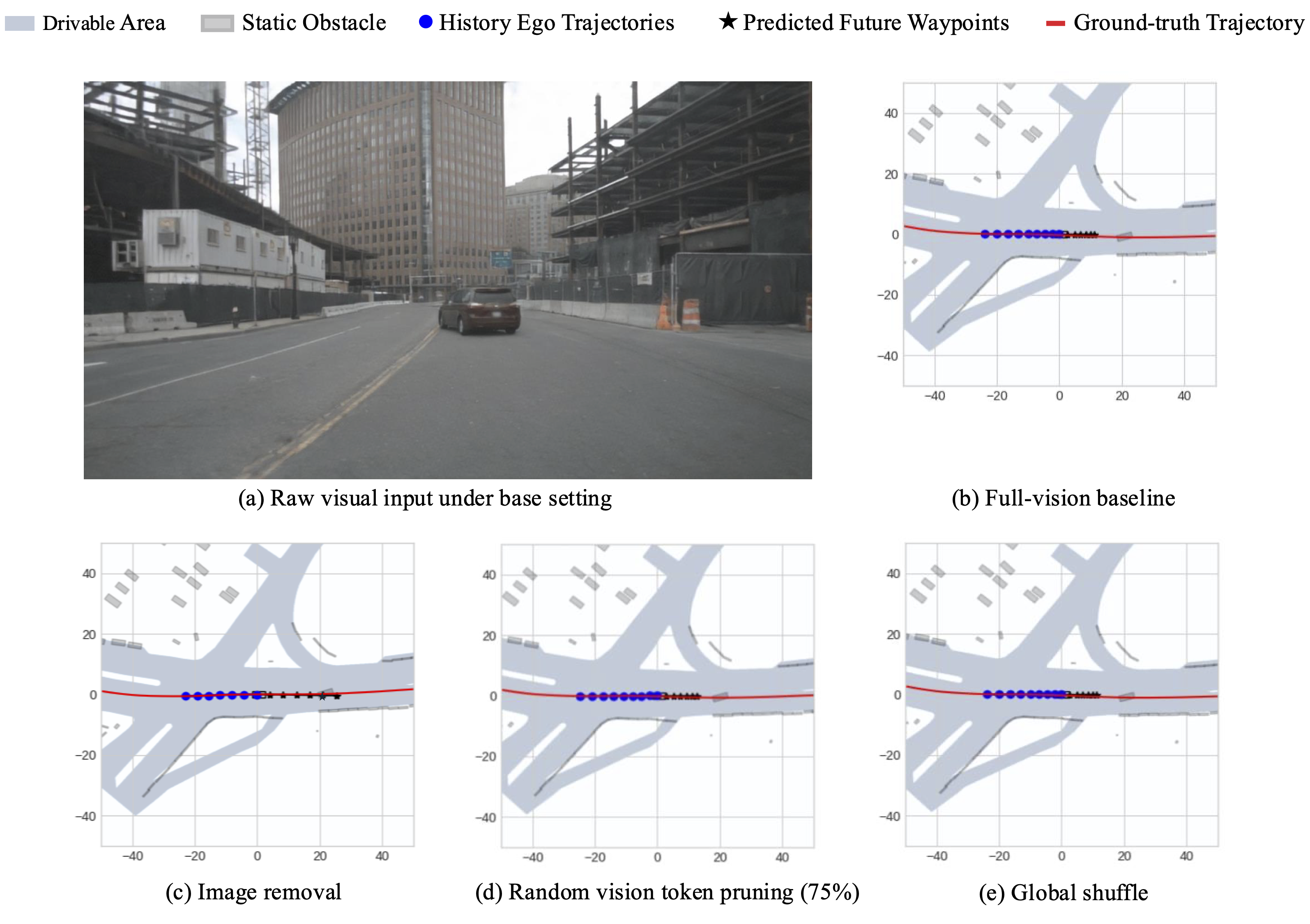}
\caption{Case study under NeuroNCAP closed-loop evaluation.
\textbf{Top:} Front-view observation and baseline behavior under full-visual input.
\label{fig:case study}
\textbf{Bottom:} BEV view under different visual perturbations.
The legend on top specifies the semantic meaning of each visual element,
including drivable area, static obstacles, historical ego trajectories, predicted waypoints, and ground-truth trajectory from the original nuScenes scenario.}

\end{figure*}

Figure~\ref{fig:case study} presents a representative example from the NeuroNCAP closed-loop benchmark to qualitatively illustrate the dependency patterns observed in quantitative evaluation. 
We analyze the \textit{stationary-106} scenario in \textit{run6} (each run uses a distinct random seed) at the 6th timestep. In this scenario, a stationary vehicle is placed along the nominal ground-truth path.
Under the full-vision setting, the vehicle begins to decelerate after detecting the stationary obstacle ahead. 
Without visual input, the model does not react to the obstacle and continues along a trajectory that remains close to the ground-truth path of the original scenario.
Random token pruning (75\%) and global shuffling both trigger deceleration, suggesting that obstacle-related cues remain reflected in the predicted behavior under these perturbations. 
Compared to global shuffling, token pruning results in weaker deceleration, leaving the vehicle closer to the obstacle at the same timestep. 
Overall, the behavioral differences under information-level and structure-level perturbations remain moderate in this example.

\section{Discussion}
The experimental results indicate that visual–behavior dependency in VLA-based driving systems varies substantially across different levels of visual perturbation.

First, visual dependency shows clear differentiation across abstraction levels. Channel-level degradation causes only moderate behavioral variation, suggesting that while the presence of visual input contributes to policy generation, low-level pixel fidelity alone is not dominant. In contrast, information-level perturbations—especially pre-encoder downsampling—lead to substantial degradation in closed-loop safety. This indicates that forming semantically meaningful visual representations is critical for stable interactive control. Meanwhile, post-encoder token pruning results in comparatively smaller performance drops, implying that planning may rely on a compact set of high-level semantic features rather than dense visual detail. Together, these findings suggest that different stages of visual processing contribute unequally to driving behavior.

Second, structure-level perturbations further highlight the importance of spatial organization. While global shuffling produces moderate degradation, positional perturbation leads to substantially larger performance drops. This contrast indicates that disrupting spatial indexing affects behavior more severely than reordering token features alone. 
The results suggest that preserving spatial consistency is important for maintaining stable cross-modal alignment and closed-loop control, which may be related to the Transformer-based architecture of the VLA model, where positional encoding and attention mechanisms organize spatial information across visual tokens.

Third, the experimental results reveal that visual dependency manifests differently under distinct evaluation settings. Open-loop trajectory metrics remain relatively stable under substantial visual perturbations, whereas closed-loop safety metrics respond more strongly to semantic and structural disruption. This contrast indicates that offline trajectory fitting and interactive safety control probe different functional roles of visual information in driving behavior.
A possible explanation is that temporal consistency from historical cues in open-loop evaluation may reduce sensitivity to certain perturbations, while interactive closed-loop scenarios require continuous perception–action feedback, where deficiencies in visual grounding are more directly reflected in safety outcomes. Importantly, this does not invalidate open-loop evaluation, but suggests that it may provide only a partial view of interaction-critical visual dependencies.

Overall, the findings suggest that visual grounding in current VLA driving systems is heterogeneous and abstraction-dependent, with different representation levels contributing unequally to behavior generation.
However, our analysis is conducted on a single representative VLA model, and extending the framework to diverse architectures and training paradigms would further validate the generality of these observations.

\section{Conclusion}
In this paper, we introduce a structured multi-level visual perturbation framework for analyzing visual–behavior dependency in VLA-based autonomous driving models. By organizing perturbations across channel-level, information-level, and structure-level dimensions, the framework enables abstraction-aware and model-agnostic diagnosis of visual grounding.
Through systematic evaluation, we demonstrate how the proposed framework enables fine-grained analysis of visual dependency across abstraction levels and evaluation contexts.
Overall, this work establishes a practical diagnostic protocol for studying visual grounding in end-to-end autonomous driving architectures and highlights the importance of explicitly analyzing visual–behavior dependency in future VLA-based systems.
\bibliographystyle{IEEEtran}
\bibliography{ref}

@misc{tang2026autodridmexplainablebenchmarkdecisionmaking,
      title={AutoDriDM: An Explainable Benchmark for Decision-Making of Vision-Language Models in Autonomous Driving}, 
      author={Zecong Tang and Zixu Wang and Yifei Wang and Weitong Lian and Tianjian Gao and Haoran Li and Tengju Ru and Lingyi Meng and Zhejun Cui and Yichen Zhu and Qi Kang and Kaixuan Wang and Yu Zhang},
      year={2026},
      eprint={2601.14702},
      archivePrefix={arXiv},
      primaryClass={cs.AI}, 
}

@article{Li2023IsES,
  title={Is Ego Status All You Need for Open-Loop End-to-End Autonomous Driving?},
  author={Zhiqi Li and Zhiding Yu and Shiyi Lan and Jiahan Li and Jan Kautz and Tong Lu and Jos{\'e} M. {\'A}lvarez},
  journal={2024 IEEE/CVF Conference on Computer Vision and Pattern Recognition (CVPR)},
  year={2023},
  pages={14864-14873}
}

@article{chi2025impromptu,
  title={Impromptu vla: Open weights and open data for driving vision-language-action models},
  author={Chi, Haohan and Gao, Huan-ang and Liu, Ziming and Liu, Jianing and Liu, Chenyu and Li, Jinwei and Yang, Kaisen and Yu, Yangcheng and Wang, Zeda and Li, Wenyi and others},
  journal={arXiv preprint arXiv:2505.23757},
  year={2025}
}

@inproceedings{caesar2020nuscenes,
  title={nuscenes: A multimodal dataset for autonomous driving},
  author={Caesar, Holger and Bankiti, Varun and Lang, Alex H and Vora, Sourabh and Liong, Venice Erin and Xu, Qiang and Krishnan, Anush and Pan, Yu and Baldan, Giancarlo and Beijbom, Oscar},
  booktitle={Proceedings of the IEEE/CVF conference on computer vision and pattern recognition},
  pages={11621--11631},
  year={2020}
}

@inproceedings{ljungbergh2024neuroncap,
  title={Neuroncap: Photorealistic closed-loop safety testing for autonomous driving},
  author={Ljungbergh, William and Tonderski, Adam and Johnander, Joakim and Caesar, Holger and {\AA}str{\"o}m, Kalle and Felsberg, Michael and Petersson, Christoffer},
  booktitle={European Conference on Computer Vision},
  pages={161--177},
  year={2024},
  organization={Springer}
}

@article{li2025recogdrive,
  title={Recogdrive: A reinforced cognitive framework for end-to-end autonomous driving},
  author={Li, Yongkang and Xiong, Kaixin and Guo, Xiangyu and Li, Fang and Yan, Sixu and Xu, Gangwei and Zhou, Lijun and Chen, Long and Sun, Haiyang and Wang, Bing and others},
  journal={arXiv preprint arXiv:2506.08052},
  year={2025}
}

@article{wang2025Alpamayo-R1,
  title={Alpamayo-r1: Bridging reasoning and action prediction for generalizable autonomous driving in the long tail},
  author={Wang, Yan and Luo, Wenjie and Bai, Junjie and Cao, Yulong and Che, Tong and Chen, Ke and Chen, Yuxiao and Diamond, Jenna and Ding, Yifan and Ding, Wenhao and others},
  journal={arXiv preprint arXiv:2511.00088},
  year={2025}
}

@article{li2025driver1,
  title={Drive-r1: Bridging reasoning and planning in vlms for autonomous driving with reinforcement learning},
  author={Li, Yue and Tian, Meng and Zhu, Dechang and Zhu, Jiangtong and Lin, Zhenyu and Xiong, Zhiwei and Zhao, Xinhai},
  journal={arXiv preprint arXiv:2506.18234},
  year={2025}
}

@article{zhou2025autovla,
  title={Autovla: A vision-language-action model for end-to-end autonomous driving with adaptive reasoning and reinforcement fine-tuning},
  author={Zhou, Zewei and Cai, Tianhui and Zhao, Seth Z and Zhang, Yun and Huang, Zhiyu and Zhou, Bolei and Ma, Jiaqi},
  journal={arXiv preprint arXiv:2506.13757},
  year={2025}
}

@article{zhou2025opendrivevla,
  title={Opendrivevla: Towards end-to-end autonomous driving with large vision language action model},
  author={Zhou, Xingcheng and Han, Xuyuan and Yang, Feng and Ma, Yunpu and Tresp, Volker and Knoll, Alois},
  journal={arXiv preprint arXiv:2503.23463},
  year={2025}
}

@article{tang2025decoupling,
  title={Decoupling Scene Perception and Ego Status: A Multi-Context Fusion Approach for Enhanced Generalization in End-to-End Autonomous Driving},
  author={Tang, Jiacheng and Feng, Mingyue and Liu, Jiachao and Wang, Yaonong and Pu, Jian},
  journal={arXiv preprint arXiv:2511.13079},
  year={2025}
}

@inproceedings{agrawal2018don,
  title={Don't just assume; look and answer: Overcoming priors for visual question answering},
  author={Agrawal, Aishwarya and Batra, Dhruv and Parikh, Devi and Kembhavi, Aniruddha},
  booktitle={Proceedings of the IEEE conference on computer vision and pattern recognition},
  pages={4971--4980},
  year={2018}
}

@article{geirhos2020shortcut,
  title={Shortcut learning in deep neural networks},
  author={Geirhos, Robert and Jacobsen, J{\"o}rn-Henrik and Michaelis, Claudio and Zemel, Richard and Brendel, Wieland and Bethge, Matthias and Wichmann, Felix A},
  journal={Nature Machine Intelligence},
  volume={2},
  number={11},
  pages={665--673},
  year={2020},
  publisher={Nature Publishing Group UK London}
}

@inproceedings{peng2022balanced,
  title={Balanced multimodal learning via on-the-fly gradient modulation},
  author={Peng, Xiaokang and Wei, Yake and Deng, Andong and Wang, Dong and Hu, Di},
  booktitle={Proceedings of the IEEE/CVF conference on computer vision and pattern recognition},
  pages={8238--8247},
  year={2022}
}

@inproceedings{goyal2017making,
  title={Making the v in vqa matter: Elevating the role of image understanding in visual question answering},
  author={Goyal, Yash and Khot, Tejas and Summers-Stay, Douglas and Batra, Dhruv and Parikh, Devi},
  booktitle={Proceedings of the IEEE conference on computer vision and pattern recognition},
  pages={6904--6913},
  year={2017}
}

@article{hendrycks2019benchmarking,
  title={Benchmarking neural network robustness to common corruptions and perturbations},
  author={Hendrycks, Dan and Dietterich, Thomas},
  journal={arXiv preprint arXiv:1903.12261},
  year={2019}
}

@inproceedings{chen2024image,
  title={An image is worth 1/2 tokens after layer 2: Plug-and-play inference acceleration for large vision-language models},
  author={Chen, Liang and Zhao, Haozhe and Liu, Tianyu and Bai, Shuai and Lin, Junyang and Zhou, Chang and Chang, Baobao},
  booktitle={European Conference on Computer Vision},
  pages={19--35},
  year={2024},
  organization={Springer}
}

@inproceedings{jiang2025survey,
  title={A survey on vision-language-action models for autonomous driving},
  author={Jiang, Sicong and Huang, Zilin and Qian, Kangan and Luo, Ziang and Zhu, Tianze and Zhong, Yang and Tang, Yihong and Kong, Menglin and Wang, Yunlong and Jiao, Siwen and others},
  booktitle={Proceedings of the IEEE/CVF International Conference on Computer Vision},
  pages={4524--4536},
  year={2025}
}

@article{sapkota2025vision,
  title={Vision-Language-Action (VLA) Models: Concepts, Progress, Applications and Challenges},
  author={Sapkota, Ranjan and Cao, Yang and Roumeliotis, Konstantinos I and Karkee, Manoj},
  journal={arXiv preprint arXiv:2505.04769},
  year={2025}
}

@InProceedings{Renz2025cvprcarllava,
  title={SimLingo: Vision-Only Closed-Loop Autonomous Driving with Language-Action Alignment},
  author={Renz, Katrin and Chen, Long and Arani, Elahe and Sinavski, Oleg},
  booktitle={Conference on Computer Vision and Pattern Recognition (CVPR)},
  year={2025}
}

@article{xie2026latentvla,
  title={LatentVLA: Efficient Vision-Language Models for Autonomous Driving via Latent Action Prediction},
  author={Xie, Chengen and Sun, Bin and Li, Tianyu and Wu, Junjie and Hao, Zhihui and Lang, XianPeng and Li, Hongyang},
  journal={arXiv preprint arXiv:2601.05611},
  year={2026}
}

@article{qwen2vl2024,
  title={Qwen2-VL: Enhancing Vision-Language Model with Advanced Visual Understanding and Multimodal Reasoning},
  author={Qwen Team},
  journal={arXiv preprint},
  year={2024}
}

@misc{gao2026steervlasteeringvisionlanguageactionmodels,
      title={SteerVLA: Steering Vision-Language-Action Models in Long-Tail Driving Scenarios}, 
      author={Tian Gao and Celine Tan and Catherine Glossop and Timothy Gao and Jiankai Sun and Kyle Stachowicz and Shirley Wu and Oier Mees and Dorsa Sadigh and Sergey Levine and Chelsea Finn},
      year={2026},
      eprint={2602.08440},
      archivePrefix={arXiv},
      primaryClass={cs.RO},
}

@inproceedings{selvaraju2019taking,
  title={Taking a hint: Leveraging explanations to make vision and language models more grounded},
  author={Selvaraju, Ramprasaath R and Lee, Stefan and Shen, Yilin and Jin, Hongxia and Ghosh, Shalini and Heck, Larry and Batra, Dhruv and Parikh, Devi},
  booktitle={Proceedings of the IEEE/CVF international conference on computer vision},
  pages={2591--2600},
  year={2019}
}

@inproceedings{niu2021counterfactual,
  title={Counterfactual vqa: A cause-effect look at language bias},
  author={Niu, Yulei and Tang, Kaihua and Zhang, Hanwang and Lu, Zhiwu and Hua, Xian-Sheng and Wen, Ji-Rong},
  booktitle={Proceedings of the IEEE/CVF conference on computer vision and pattern recognition},
  pages={12700--12710},
  year={2021}
}

@inproceedings{dong2023benchmarking,
  title={Benchmarking robustness of 3d object detection to common corruptions},
  author={Dong, Yinpeng and Kang, Caixin and Zhang, Jinlai and Zhu, Zijian and Wang, Yikai and Yang, Xiao and Su, Hang and Wei, Xingxing and Zhu, Jun},
  booktitle={Proceedings of the IEEE/CVF Conference on Computer Vision and Pattern Recognition},
  pages={1022--1032},
  year={2023}
}

@INPROCEEDINGS{10733168,
  author={Qiu, Yongsheng and Lu, Yuanyao and Wang, Yuantao and Yang, Chaochao},
  booktitle={2024 IEEE 7th Information Technology, Networking, Electronic and Automation Control Conference (ITNEC)}, 
  title={Visual Perception Challenges in Adverse Weather for Autonomous Vehicles: A Review of Rain and Fog Impacts}, 
  year={2024},
  volume={7},
  number={},
  pages={1342-1348},
  doi={10.1109/ITNEC60942.2024.10733168}}
\end{document}